\documentclass{article} 
\usepackage{colm2024_conference}

\usepackage{microtype}
\usepackage{hyperref}
\usepackage{url}
\usepackage{graphicx}
\usepackage{booktabs} 
\usepackage{multirow}
\usepackage{marvosym}
\usepackage{amsmath}
\usepackage{amssymb} 
\usepackage{cleveref}

\title{Do We Really Need a Complex Agent System? Distill Embodied Agent into a Single Model}

\author{Zhonghan Zhao$^{1}$, Ke Ma$^{1}$, Wenhao Chai$^{2,\dagger}$, Xuan Wang$^{1}$, Kewei Chen$^{3}$ \\
\textbf{Dongxu Guo$^{3}$, Yanting Zhang$^{3}$, Hongwei Wang$^{1}$ and Gaoang Wang$^{1,\text{\Letter}}$}\\
$^{1}$ Zhejiang University \quad
$^{2}$ University of Washington \quad
$^{3}$ Donghua University \\
\texttt{\{zhonghan.22, gaoangwang\}@intl.zju.edu.cn, wchai@uw.edu}
}

\newcommand\lft{\mathopen{}\left}
\newcommand\rgt{\aftergroup\mathclose\aftergroup{\aftergroup}\right}

\colmfinalcopy 
\begin{document}

\maketitle
\renewcommand{\thefootnote}{\fnsymbol{footnote}}
\footnotetext{\textsuperscript{$\dagger$}Project Lead, \textsuperscript{\Letter} Corresponding author.}
\renewcommand*{\thefootnote}{\arabic{footnote}}

\begin{abstract}
With the power of large language models (LLMs), open-ended embodied agents can flexibly understand human instructions, generate interpretable guidance strategies, and output executable actions. Nowadays, Multi-modal Language Models~(MLMs) integrate multi-modal signals into LLMs, further bringing richer perception to entity agents and allowing embodied agents to perceive world-understanding tasks more delicately. 
However, existing works: 1) operate independently by agents, each containing multiple LLMs, from perception to action, resulting in gaps between complex tasks and execution; 2) train MLMs on static data, struggling with dynamics in open-ended scenarios; 3) input prior knowledge directly as prompts, suppressing application flexibility. 
We propose STEVE-2, a hierarchical knowledge distillation framework for open-ended embodied tasks, characterized by 1) a hierarchical system for multi-granular task division, 2) a mirrored distillation method for parallel simulation data, and 3) an extra expert model for bringing additional knowledge into parallel simulation. After distillation, embodied agents can complete complex, open-ended tasks without additional expert guidance, utilizing the performance and knowledge of a versatile MLM.
Extensive evaluations on navigation and creation tasks highlight the superior performance of STEVE-2 in open-ended tasks, with $1.4 \times$ - $7.3 \times$ in performance.
\end{abstract}
\section{Introduction}
Drawing on large language models (LLMs) with human knowledge, communicative competence, and decision-making capabilities, embodied agents exhibit human-like intelligence in simulators~\citep{park2023generative, wang2023voyager, zhao2023see}, programming~\citep{qian2023experiential, hong2023metagpt}, and robotic tasks~\citep{zhang2023building, mandi2023roco}. The development of individual intelligence has led to the creation a new, collaborative framework where multiple agents~\citep{chen2023autoagents, chen2023agentverse} work together. Agents in this system specialize in complex tasks, sharing insights to enhance efficiency and foster a learning environment. 

For open-ended environments, the need to solve complex tasks makes people naturally turn to multi-agent cooperation. However, current methods~\citep{wang2023voyager, zhao2023see} infers based on a combination of multiple LLMs and prompts with different functions to achieve sufficient performance in single-agent tasks. This approach incurs high reasoning costs and seriously hinders the upper limit of the development of open-ended multi-agents~\citep{zhao2024hierarchical}. Agents require more prior knowledge to understand complex task descriptions. Creative agents~\citep{zhang2023creative} use images generated by a model to help construct and understand goals, but this can be costly and limits flexibility.

LLM-based agents interact with text-based environments and are skilled in planning, reflection, and reward shaping. They achieve this through their reasoning capability and semantic abstraction. However, LLMs cannot be utilized in such settings due to the complexity of visual environments. While some researchers~\citep{zheng2023steve, zhao2023see} have tried to integrate visual encoders into LLMs, the performance obtained by training on static data using offline training cannot match the dynamics of the environment. The challenge is fine-tuning Multi-modal Language Models~(MLMs) as an embodied agent with dynamic alignment to the visual world. EMMA~\citep{yang2023embodied} provides a solution by training an embodied reflex agent through aligning an MLM with visual world dynamics and distilling API LLM skills. Using the DPO loss~\citep{rafailov2024direct} as a training method is highly effective and also serves as a knowledge distillation technique. By combining the teacher model with the extra expert, we can integrate additional expert knowledge into the model, which is crucial for completing complex, open-ended tasks. We also observe that the fine-grained Chain-of-Thoughts~(CoT)~\citep{wei2022chain} is necessary for improving the reasoning accuracy of LLMs on complex tasks. Evoked from Centralized Training with Decentralized Execution~(CTDE)~\citep{hu2023attention,sunehag2017value,lowe2017multi,mao2018modelling} framework for MARL, we propose Centralized Planning with Decentralized Execution~(CPDE) a hierarchical structure maximizing global oversight and local autonomy and simulate refined CoT to achieve multi-granular task division and fine-grained action.

As shown in~\Cref{fig:introduction}, we propose \textbf{STEVE-2} a hierarchical knowledge distillation process employing DPO loss~\citep{rafailov2024direct} on Multi-Agent System with Extra Expert module. After distillation, \textbf{STEVE-2} can develop efficient embodied agents to accomplish fine-grained open-ended tasks without expert guidance with \textbf{one model}. 
We summarize our contributions as follows:

\begin{figure*}[t]
\centering
    \includegraphics[width=\linewidth]{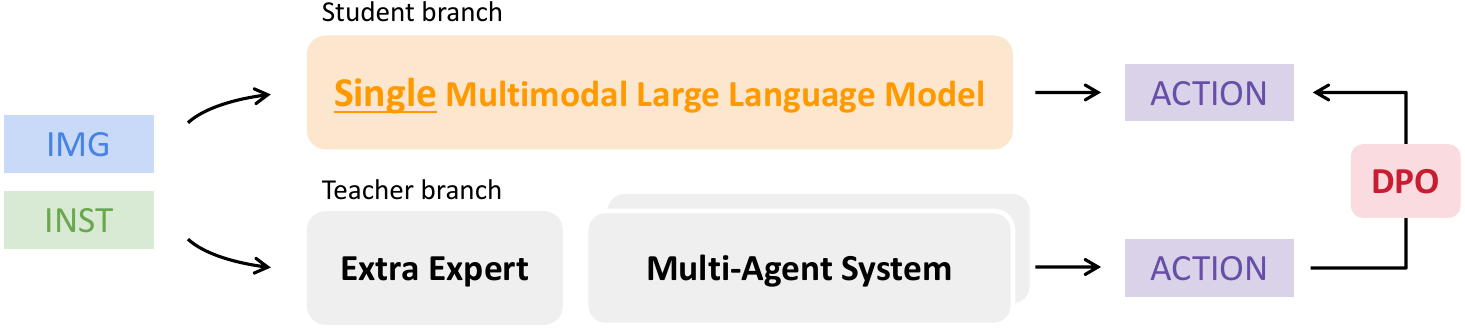}\\
    \caption{\small \textbf{STEVE-2 trained via knowledge distillation.} Based on DPO loss~\citep{rafailov2024direct}, \textbf{STEVE-2} is distilled from the Teacher Model, parallelly working with the same input, image, and text instruction. Extra Expert is an extra model for clarification of goals. IMG and INST are image and text instructions.
    }
    \label{fig:introduction}
\end{figure*}

\noindent \textbf{$\bullet$} 
We design \textbf{STEVE-2}, hierarchical knowledge distillation for multi-modal multi-agent training. Multi-agents cooperate in a hierarchical auto-organizing system for fine-grained Chain-of-Thoughts and efficient deployment. Each agent is trained hierarchically by mirrored teachers for simulating dynamics and aligning tasks at multiple levels of granularity, allowing efficient cooperation using only \textbf{one MLM}.

\noindent \textbf{$\bullet$} 
We develop an extra expert model based on our distillation method, which allows us to add extra multi-modal knowledge to parallel teachers and implicitly transfer knowledge to the inference model. By keeping the knowledge within the model, there is no need for additional priori or expert guidance during reasoning, and the model can handle multi-agent cooperation flexibly and efficiently.

\noindent \textbf{$\bullet$} 
We achieve state-of-the-art performance on the asynchronous multi-modal navigation and creation tasks in Minecraft's open-ended environment, with $1.4 \times$ - $7.3 \times$ in performance.

\section{Related Works}

\subsection{Embodied Multimodal Model}

Embodied agents integrate sensory perceptions, physical actions, and computational intelligence to accomplish tasks and goals within their environment. Key areas are wide-ranging, including Navigation~\citep{wijmans2019dd,yu2021sound,du2020vtnet,kwon2023renderable,chen2021history,moudgil2021soat}, Embodied Question Answering~\citep{das2018embodied,datta2022episodic}, Active Visual Tracking~\citep{luo2018end,zhong2021towards,luo2019end,zhong2019ad}, and Visual Exploration~\citep{liu2022symmetry,dean2020see,chen2018learning}. The field is evolving rapidly with the development of Large Language Models (LLMs)~\citep{song2022llm} and Multimodal LLMs (MLLMs)~\citep{alayrac2022flamingo,zhu2023minigpt,li2023otter,li2022blip,li2023blip,gong2023multimodal,lyu2023macaw,ye2023mplug,dai2023instructblip,wang2023visionllm,liu2023visual,maaz2023video,su2023pandagpt,gao2023llama}, integrating multiple modalities for more effective processing. A prime example of this innovation is PaLM-E~\citep{driess2023palm}, a sophisticated multimodal model with 562B parameters, adept at a broad spectrum of embodied tasks and demonstrating exceptional capabilities in visual reasoning.

\subsection{LLM-based Multi-Agent Frameworks} 

Large Language Models~(LLMs) are skilled at completing new tasks when given prompt-based instructions. Autonomous agents based on Large Language Model-based~(LLM-based) models have gained significant interest in industry and academia~\citep{wang2023survey}. 
Several works~\citep{wang2023unleashing,du2023improving,zhuge2023mindstorms,hao2023chatllm,akata2023playing} have augmented the problem-solving abilities of LLMs by incorporating discussions among multiple agents. Stable-Alignment~\citep{liu2023training} generates instruction datasets by reaching a consensus on value judgments through interactions among LLM agents in a sandbox. Some works in the field of artificial intelligence focus on studying sociological phenomena. For instance, Generative Agents~\citep{park2023generative} creates a virtual ``town'' comprising 25 agents to investigate language interaction, social understanding, and collective memory. The Natural Language-Based Society of Mind~\citep{zhuge2023mindstorms} involves agents with different functions interacting to solve complex tasks through multiple rounds of Mindstorms. In addition, others~\citep{cai2023large} propose a model for cost reduction by combining large models as tool makers and small models as tool users.
Some works emphasize cooperation and competition related to planning and strategy~\citep{meta2022human}, some propose LLM-based economies~\citep{zhuge2023mindstorms}, and others propose LLM-based programming~\citep{hong2023metagpt}. The current method uses many language models~\citep{wang2023voyager}, making it slow and inefficient, particularly in open-ended environments. Control capabilities for LLM agents~\citep{zheng2023steve} are challenging, but this is what Reinforcement Learning methods~\citep{lifshitz2023steve, cai2023groot, cai2023open} do best. However, combining LLM with Reinforcement Learning models~\citep{wang2023jarvis} can improve control capabilities but lacks flexibility. To improve multi-agent cooperation, we aim to reduce the number of language models and increase efficiency through a hierarchical organization system and knowledge distillation while ensuring control capabilities.

\section{STEVE-2: Hierarchical Knowledge Distillation}

\begin{figure*}[t]
\centering
    \includegraphics[width=\linewidth]{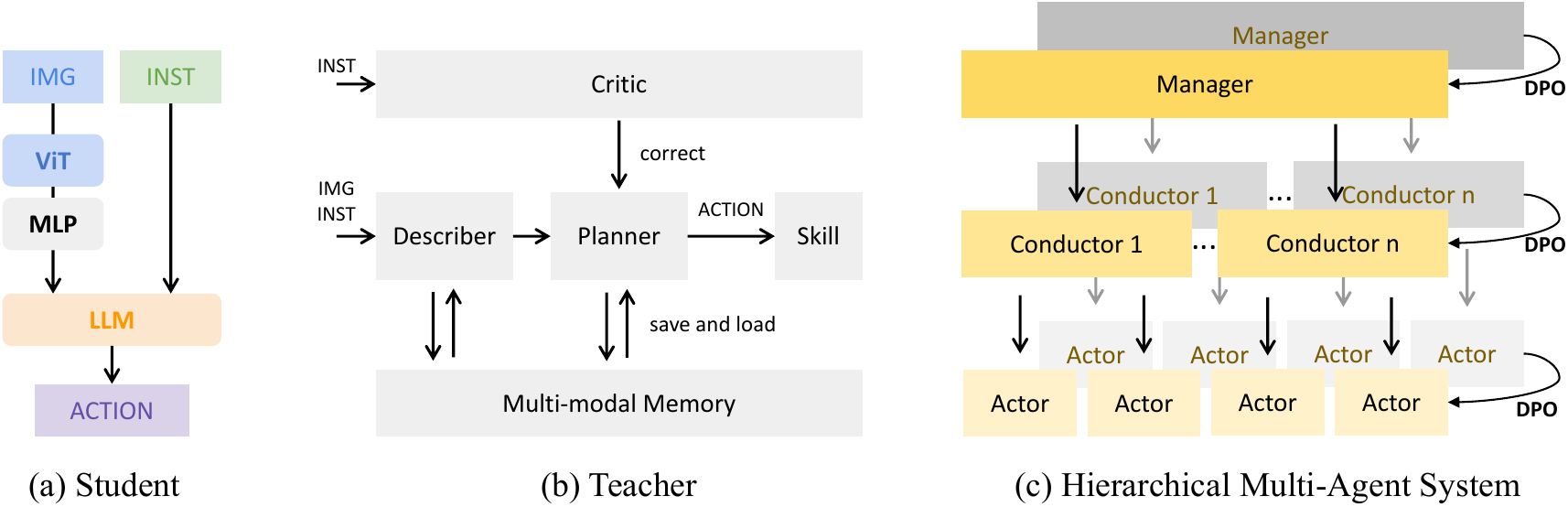}\\
    \caption{\small \textbf{STEVE-2 framework.} (a) STEVE-2 Agent takes a textual task description and pixel observations as its input state per step to generate a sequence of actions knowledge distilled from (b) Teacher Agent, which is a combination of multi-functional MLM. Both work parallel in (c) Multi-Agent System, a hierarchical organizational structure: grey for the teacher agent and yellow for the steve-2 agent.
    }
    \label{fig:framework}
\end{figure*}

As shown in~\Cref{fig:framework}, the \textbf{STEVE-2} is a hierarchical MLM-based multi-agent system denoted as $\mathcal{F}$, which can manage and execute complex multi-agent tasks $\mathcal{T}$ on visual~$v$, audio~$a$, and object~$o$ goals with perception on the state list of vision~$v$, audio~$a$, and other properties~$p$ within open-ended environments by leveraging cognitive and collaborative capabilities of the multi-modal language model:
\begin{equation}
\mathcal{S} = \{(s^v_i, s^a_i, s^p_i)\}_{i\leq N},\\\ \mathcal{T} = (t^v_i, t^a_i, t^o_i), 
\end{equation}
where $\mathcal{S}_i = (s^v_i, s^a_i, s^p_i)$ represents the state list of vision, audio, and other properties of one conductor agent, $\mathcal{T}$ is the initial task. Then, we get \( \mathbf{a}^c_i\) as the action for the conductor agent \( i\), $N$ is the number of conductor agents.

The Hierarchical architecture consists of two primary operational domains: higher-order centralized planning, which is managed by the manager multi-modal language model ($\mathbf{MLM}^M$), and ground-level decentralized execution, which is conducted by the conductor model ($\mathbf{MLM}^C$), then the action $a^c_i$ of the conductor can be obtained as follows,
\begin{equation}
\mathbf{a}^c_i = \mathcal{F}(\mathcal{S},\mathcal{T}) = \mathbf{MLM}^C(\mathbf{MLM}^M(\mathcal{S},\mathcal{T}), \mathcal{S}_i),
\label{eqn:main_function}
\end{equation}
where $\mathbf{MLM}^C$ and $\mathbf{MLM}^M$ represent the multi-modal language models of the conductor and manager agents. Actor agents $\mathbf{MLM}^A$ auto-organized by $\mathbf{MLM}^C$ are optional for additional actions: $\mathbf{a}^a = \mathbf{MLM}^A(\mathbf{MLM}^C(\cdot))$. To achieve this, we draw inspiration from recent advances of large pretrained MLMs~\citep{liu2023improvedllava,liu2023llava}. 
As shown in~\Cref{fig:framework}, each teacher agent plays the above three different MLMs through different prompts in the hierarchical multi-agent system. 
After hierarchical knowledge distillation by DPO loss~\citep{rafailov2024direct}, \textbf{STEVE-2} acquires the performance of these three scale multi-modal language models $\mathbf{MLM} = \{\mathbf{MLM}^M, \mathbf{MLM}^C, \mathbf{MLM}^A\}$ for the manager, conductor, and actor agents. \textbf{STEVE-2} is modularized into three components: 

\begin{itemize}
    \item [$\bullet$] \textbf{The pretrained Vision Transformer (ViT)} acts as the vision encoder that encodes visual input data (represented as $s^v$) into visual embeddings. 
    \item [$\bullet$] \textbf{The Multi-Layer Perceptron (MLP) layer} aligns the embeddings produced by the ViT with the language space. 
    \item [$\bullet$] \textbf{The pretrained Language Model (LLM)} is utilized as the language decoder, taking the concatenation of instruction tokens and the output of the linear projection layer as input and generates a textual action $\mathbf{a}^t$. This textual action is then used to retrieve code action $\mathbf{a}^c$.
\end{itemize}

\subsection{Multi-modal Teacher Model} \label{sec:MLM}
The Multi-modal Teacher Model $\mathbf{MLM} = \{\mathbf{MLM}^M, \mathbf{MLM}^C, \mathbf{MLM}^A\}$, consisting of three different types of MLM of the manager, conductor and actor agents. As shown in~\Cref{fig:framework}, $\mathbf{MLM}^M, \mathbf{MLM}^C, \mathbf{MLM}^A = \{\mathcal{P}_l, \mathcal{D}_s, \mathcal{C}_r, \mathcal{S}_k\}$, which consists of Planner, Describer, Critic, and  Skill module for the manager, conductor and actor agents. They formulate aligned task plans, condense and translate multi-modal data, refine strategies through feedback, and assign and direct agent subtasks. Then, they translate strategic plans into executable actions, orchestrate dynamic group formations, and distribute tasks across agents, ensuring alignment with centralized directives and facilitating continuous learning and adaptation through a curriculum of complex tasks. A multi-modal memory is maintained to store the long-term memory of the description of multi-modal information generated by $\mathcal{D}_s$, the specific planning trajectories generated by $\mathcal{P}_l$ corrected by $\mathcal{P}_l$.

\paragraph{Adaptive planning with MLM.}
Our approach integrates the environment's observations and task directives to plan actions based on the current scenario. We begin by translating multi-modal observations into textual descriptions, utilizing a method that avoids direct scene captioning by the MLM. Instead, we extract keywords for items from the STEVE-21K dataset~\citep{zhao2023see} and employ GPT-4V(ision)~\citep{yang2023dawn} to craft sentences that articulate these observations. The MLM identifies relevant condition sentences from textual observations during the planning phase. It also incorporates additional context, such as biome types and inventory levels, into text formats via predefined templates. We generate action plans by re-engaging the MLM's linguistic component with the task instructions and these descriptive texts. This methodology leverages the MLM's capabilities in a layered manner, yielding more accurate situational descriptions and plans that significantly reduce the likelihood of generating unrealistic elements compared to fully integrated models.

\paragraph{Autonomous error correction and proactive planning.}

\textbf{STEVE-2} enhances its planning through a closed-loop feedback mechanism, automatically correcting failures by analyzing feedback and identifying errors using its self-explanation capabilities. Unlike other agents, it generates improved plans without human input or extra information. Additionally, \textbf{STEVE-2} simulates and evaluates each plan step to identify potential flaws early, reducing the likelihood of encountering difficult situations due to plan failures. This proactive approach enables it to foresee issues like insufficient resources, which could hinder task completion.

\subsection{Teacher with Extra Expert}
Language instructions can be limited in their ability to describe complicated tasks due to the abstract nature of language. This can lead to high uncertainty in task completion and may even require the agent to possess creativity. Although many open-ended agents~\citep{lifshitz2023steve, zhao2023see} can follow clear instructions for specific task goals, none can follow uncertain instructions to perform complicated tasks. Navigation and creation tasks are particularly challenging as they are open-ended, and the language instructions provided may lack information and refer to diverse, complicated outcomes. This requires the agent to imagine details not specified in the instructions. A simple instruction like ``Find a Pyramid'', or ``Build a Pogda'' can refer to a complex task, making planning and execution difficult. We have taken inspiration from the work of Creative Agents~\citep{zhang2023creative}, which uses a generation model as the prior input to simulate the imagination of a house image, represented by the imaginator $\mathcal{I}$.
However, restoring other perspectives and internal structures solely through a 2D image from one perspective is challenging when creating a building. For navigation tasks, having only one concrete image from the imaginator will be difficult for long-distance searches. To address this issue, we have extended $\mathcal{I}$ to the extra $\mathbf{Expert}$, modified VQ-VAE~\citep{razavi2019generating} that generates 3D occupancy for creation task, and a dynamic map~\citep{zhao2024hierarchical} with multi-modal signals for navigation task. Note that we have integrated this part only with the teacher model to avoid providing direct prompt information to the \textbf{STEVE-2} part, which can lead to cheating. This approach also ensures the lightweight of the \textbf{STEVE-2}.
The teacher with extra expert is shown below:
\begin{equation}
    P(\mathbf{a}^t|\mathcal{S},t^o) = \sum_g{\mathbf{Expert}(g|t^o)\mathcal{F}_T(\mathbf{a}^t|\mathcal{S}, t^o, g)},
\end{equation}
where $g \in G$ is the multi-modal knowledge of the textual description. The state is represented by $\mathcal{S}$. $\mathbf{Expert}$ converts the textual object goal $t^o \in \mathcal{T}$ into $g$. This process provides a detailed task description to the teacher model $\mathcal{F}_T$. This approach allows us to develop more advanced agents to handle creative tasks by utilizing the extra expert's ability to generate diverse outcomes. This provides the agent with richer task information, reducing uncertainty.

\subsection{Knowledge Distillation }
Our goal is to develop a system of hierarchical MLM agents, and each assigned a specific function represented by a student MLM $\mathcal{F}_{\theta}$, that can learn and replicate the behavior of teacher MLMs $\mathcal{F}_T(\cdot)$ with extra knowledge: $\mathbf{a} = \mathcal{F}_\theta(\cdot)$ and $\mathbf{a}^{*} = \mathcal{F}_T(\cdot)$. We employ a similar approach to our system in training the Manager, Conductor, and Actor agents in \textbf{STEVE-2} using a hierarchical MLM teacher. The objective function, which is based on the state distribution induced by $\mathcal{F}_{\theta}$, is minimized to accomplish this.
\begin{equation}
    \theta^{*} = \mathop{\arg\min}\limits_{\theta\in\Theta}\mathcal{L}_{\text{KD}}(\mathcal{F}_{\theta}(\mathbf{a}|\mathcal{S}),\mathbf{a}^{*})),
    \label{eq:kd_loss}
\end{equation}
where the loss function $\mathcal{L}_{\text{KD}}$ is the knowledge distillation loss. Specifically, we adopt DPO loss~\citep{rafailov2024direct} as the distillation loss in the experiment. It uses the relative log probability of a response to the non-preferred response, with a dynamic per-sample weight to prevent model degradation, and is proven better than cross-entropy for aligning language models with teacher preferences. Therefore,~\Cref{eq:kd_loss} can be extended to the following formulation:
\begin{align}
    &\theta^{*} = \mathop{\arg\min}\limits_{\theta\in\Theta}\mathcal{L}_{\text{DPO}}(\mathcal{F}_{\theta},\mathcal{F}_{\text{ref}},\mathcal{S},\mathbf{a},\mathbf{a}^{*}), \label{eq:dpo}\\ 
    &\mathcal{L}_{\text{DPO}}(\cdot) \triangleq -\mathbb{E}_{\mathcal{F}_{\theta}}\lft[\log\sigma\lft(\beta\log\frac{\mathcal{F}_{\theta}(\mathbf{a}^{*}|\mathcal{S})}{\mathcal{F}_{\text{ref}}(\mathbf{a}^{*}|\mathcal{S})} - \beta\log\frac{\mathcal{F}_{\theta}(\mathbf{a}|\mathcal{S})}{\mathcal{F}_{\text{ref}}(\mathbf{a}|\mathcal{S})}\rgt)\rgt], 
\end{align}
where a logistic function ($\sigma$) and a hyperparameter ($\beta$) are to control the deviation from a reference agent $\mathcal{F}_{\text{ref}}$. The reference agent is obtained by behavior cloning on a dataset produced by a rule-based teacher. The model is trained using Maximum Likelihood Estimation (MLE) with an additional regularization term to prevent the agent from deviating too far from the teacher's accurate distribution, maintain generation diversity, and avoid premature convergence to easy tasks. To stabilize the training process, the MLM agent $\mathcal{F}_{\theta}$ is initialized to $\mathcal{F}_{\text{ref}}$ as well. However, since we cannot compute the distribution of $\mathcal{F}_{\text{ref}}$, we must use the agent to sample $\mathcal{F}_{\theta}$. We used DAgger~\citep{sun2023mega} to converge to the optimal agent $\mathcal{F}_{\theta^{*}}$ and overcome cumulative error and distribution shift issues instead of Naive behavior. Note that we use GPT-4V(ision)~\citep{yang2023dawn} to generate a sequence of actions, represented by $\mathbf{a}^{*}$ in equation DPO~\citep{rafailov2024direct} for a given task. With extra expert, these actions may be closer to optimal.
\section{Experiment}
We experiment on two tasks to evaluate multi-agent cooperation and running efficiency on complex tasks. We discuss evaluation settings, results, and a trial on long-horizon tasks. 

\subsection{Experimental Setups}
Our STEVE-2 is based on STEVE-13b~\citep{zhao2023see} pretrained on LLaMA2-13b~\citep{touvron2023llama}, and we select gpt-4-1106-vision-preview~\citep{yang2023dawn} as the teacher model. Our simulation environment is based on MineDojo~\citep{fan2022minedojo} and uses Mineflayer~\citep{mineflayer} APIs for motor controls. The generative model is a 3D occupancy generator that aligns text input based on the VQ-VAE~\citep{razavi2019generating}. The dynamic map~\citep{zhao2024hierarchical} is a multi-modal memory updated in real-time with information from all action agents, providing a strategic overview of the environment. The maximum number of robots that can be allocated based on this environment is 8, which is also our experimental robots' upper limit. 

\subsection{Baselines}
No pre-built multi-agent robots for Minecraft exist. So, we choose representative algorithms as baselines for our experiment. These algorithms extract information from the system's backend, differing from real-world scenarios.

\paragraph{Voyager}\citep{wang2023voyager} is a blind, single-robot system relying only on textual grounding for perception. It has a long-term procedural memory that stores a hierarchical library of code-based grounding procedures. Voyager is known for its ability to explore areas and master the tech tree. However, its main focus is to prompt GPT-4~\citep{openai2023gpt4} on background text messages in embodied agents. We use multiple hosts to deploy multiple models to work directly on the server. We convert the input of the image goal into a text task through GPT-4V(ision)~\citep{yang2023dawn}, using the same Describer module as our teacher model.

\paragraph{STEVE}\citep{zhao2023see} is a multi-modal single-robot system that combines the vision unit with the STEVE-13B with the code database. It focuses on introducing a visual module to endow the model with visual perception capabilities in processing visual perception information and handling task reasoning for skill-related execution. Similarly, using multiple hosts, we deploy multiple models to work directly on the server.

\paragraph{Creative Agents}\citep{zhang2023creative} is a blind, single-robot system enhanced with a diffusion model that generates detailed imaginations of task outcomes conditioned on language instructions for open-ended creative tasks. They observed that diffusion with GPT-4V(vision)~\citep{yang2023dawn} performs best in the creation task. Hence, we use this version and test replacing diffusion with our finetuned VQ-VAE~\cite{razavi2019generating} to provide image input.

\subsection{Task \uppercase\expandafter{\romannumeral1}: Multi-modal Navigation}
\begin{table}[!t]
\centering
\resizebox{\linewidth}{!}{
\begin{tabular}{@{}l|c|cc|cc|cc@{}}
\toprule
\multirow{2}{*}{Method} & \multirow{2}{*}{\# LLMs} & \multicolumn{2}{c|}{\bf Goal Search} & \multicolumn{2}{c|}{\bf Block Search} & \multicolumn{2}{c}{\bf Map Explore}\\
\cmidrule(lr){3-4} \cmidrule(lr){5-6} \cmidrule(lr){7-8}
& & \# iters~($\downarrow$) & success~($\uparrow$) & \# iters~($\downarrow$) & \# blocks~($\uparrow$) & \# iters~($\downarrow$) & \# area~($\uparrow$)  \\
\midrule
\multirow{2}{*}{Voyager} & 4 & - & 43\% & 34 & 28 & 6 & 175 \\
                 & 12 / 12 / 20 & 29 & 64\% & 14 & 81 & 3 & 755 \\
\midrule
\multirow{2}{*}{STEVE} & 4 & 56 & 42\% & 15 & 65 & 6 & 161 \\
              & 20 / 16 / 24  & 22 & 64\% & 6 & 211 & 3 & 696 \\ 
\midrule
\multirow{2}{*}{\bf STEVE-2} & 1 & 15 & 66\% & 12 & 73 & 6 & 170\\
                        & 5 / 8 / 8 & \textbf{4}  & \textbf{91\%} & \textbf{3} & \textbf{412} & \textbf{1} & \textbf{1493}\\
\bottomrule
\end{tabular}
}
\caption{\textbf{Comparison on multi-modal navigation task.} \# iters represent the average number of prompting iterations required to finish each task, with a maximum of 5 for map exploration and 100 iterations for others. Success rate is for task fulfillment. \# blocks denotes the average number of diamond blocks found over 100 iterations. \# area denotes the average squares of blocks over 5 iterations. We list the one-agent and best performance with the number of LLMs~(from left to right represents different goals) on the bottom line.}
\label{tab:navigation}
\end{table}

\vspace{-10pt}

We conduct experiments on navigation tasks, including multi-modal goal search, continuous block search, and map exploration. Our \textbf{STEVE-2} achieves the best performance, indicating the high efficiency of multi-agent cooperation in multiple navigation tasks with different modes of operation. The results are shown in~\Cref{tab:navigation}. Process efficiency and module reduction are greatly improved when fewer LLMs are used.

\paragraph{Multi-modal goal search.}
Multi-modal goal search technique identifies Image, Object, and Audio goals. Object labels identify in-game items, Image labels locate objects using images, and Audio labels detect sounds outside the player's range. Our multi-agent system, \textbf{STEVE-2}, decomposes tasks layer by layer and achieves 5.5$ \times$ better performance than the state-of-the-art LLM-based method while maintaining the same success rate.

\paragraph{Continuous block search.}
Continuous block search is to assess an agent's exploratory abilities and proficiency in locating diamond blocks. The agent will perform a block searching task, identifying as many blocks as possible in the fewest iterations. The dynamic map and hierarchical structure will allow us to find and deploy more blocks. We experiment with the same setting as detailed in the paper~\citep{zhao2023see}.

\paragraph{Map exploration.}
Map exploration aims to let the agent update the map as much as possible. We set up the same status awareness: when in an unreached area, status information prompts the agent in text. We set each step's maximum movement distance not to exceed 50 blocks. 
STEVE-2 achieved 1.9 $\times$ better performance than the state-of-the-art LLM-based method while increasing efficiency by 3 $\times$.

\begin{figure*}[t]
\centering
    \includegraphics[width=\linewidth]{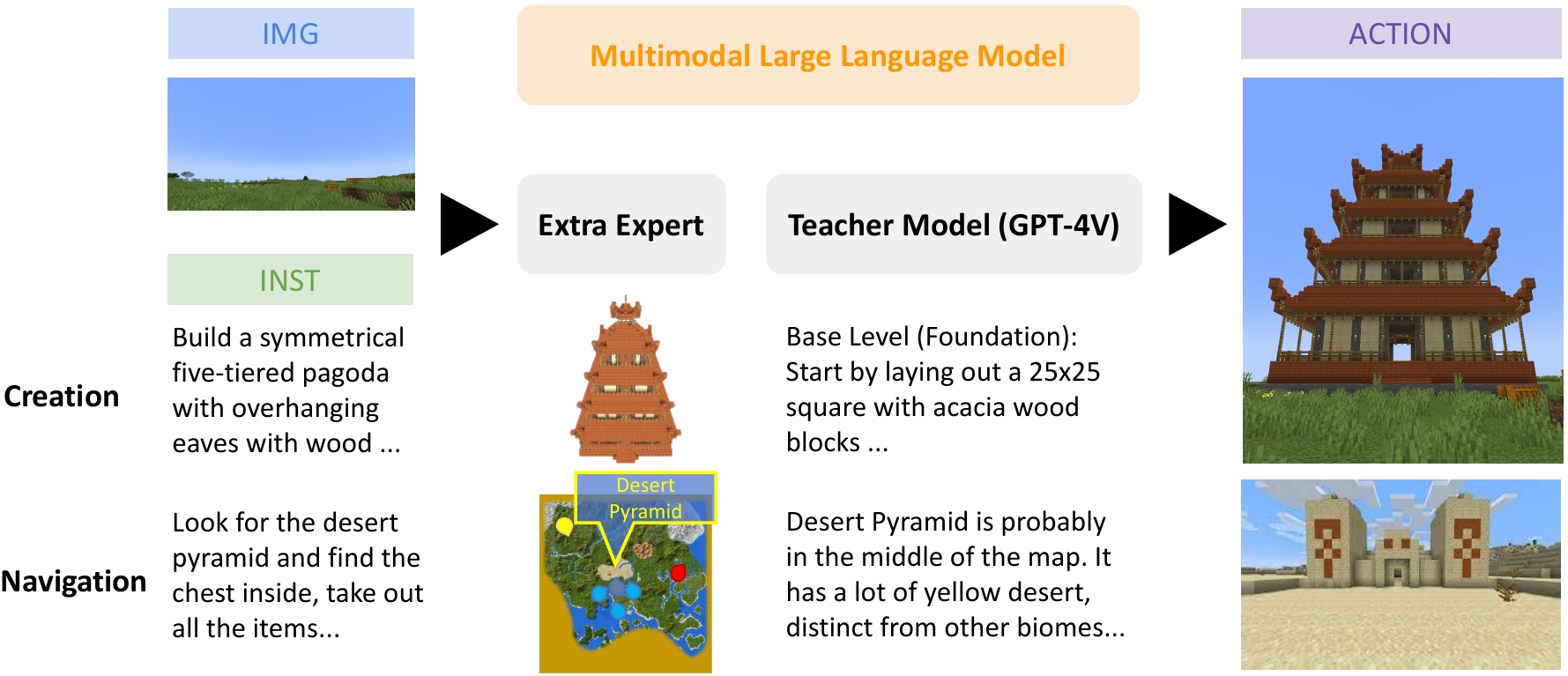}\\
    \caption{\small \textbf{Overview of multi-modal creation and navigation.} Extra Expert comprises the generative model and dynamic map for both tasks. The generative model generates 3D occupancy, rendered into multi-view images and input to the Teacher Model. 
    }
    \label{fig:illustration}
\end{figure*}

\subsection{Task \uppercase\expandafter{\romannumeral2}: Multi-modal Creation}
\begin{table}[!t]
\centering
\resizebox{1.0\linewidth}{!}{
\begin{tabular}{@{}l|c|cc|cc@{}}
\toprule
\multirow{2}{*}{\bf Setting} & \multirow{2}{*}{\# LLMs} & \multicolumn{2}{c|}{\bf Material Collection} & \multicolumn{2}{c}{\bf Building Creation}\\
\cmidrule(lr){3-4} \cmidrule(lr){5-6}
& & \# iters~($\downarrow$) & completion~($\uparrow$) & FID~($\downarrow$) & Preferance~($\uparrow$)\\
\midrule
\multirow{1}{*}{Voyager} & 4 & 76 & 72\% & 256.75 & 3.6\\
\midrule
\multirow{1}{*}{Creative Agents} & 4 & - & - & 68.32 & 6.7 \\
\midrule
\multirow{2}{*}{\bf STEVE-2~(Ours)} & 1 & 41 & 95\% & 67.38 & 5.5\\
                        & 8 / 2 & \textbf{4}  & \textbf{99\%} & \textbf{21.12} & \textbf{7.7}\\
\bottomrule
\end{tabular}
}
\caption{\textbf{Comparison on multi-modal creation task.} The creation task consists of material collection and building creation. \# iters represent the average number of iterations required to finish the task with a maximum of 100 prompting iterations. Completion rate is the completion rate. FID is the average FID score of multi-perspective images. Preference score is the average of human and GPT-4V scores, rated from 0-10. 
}
\label{tab:creation}
\end{table}

\vspace{-10pt}
To evaluate how well the creation task works, we divided it into two parts: collecting materials and building. This approach helped us assess the team's efficiency and ability to understand complex task instructions and translate them into precise actions.
\textbf{STEVE-2} has achieved the best performance, reflecting teamwork efficiency and understanding abstract task descriptions. For more details of case study, please refer to~\cref{sec:case}.

\vspace{-10pt}
\paragraph{Material collection.}
Material collection is vital for architecture and multi-agent cooperation. We'll give the agent a list of materials and their quantities to collect efficiently. We test only Voyager, as creative agents are based on Voyager.
\textbf{STEVE-2} can improve efficiency by 19 $\times$ while ensuring accuracy.

\vspace{-10pt}
\paragraph{Building creation.}
Building creation is challenging due to novel and complicated outcomes, requiring the agent to imagine details unspecified by the instruction.
The finetuned VQ-VAE~\citep{razavi2019generating} modified to 3D occupancy generation is used as the imagination, while multiple perspectives of occupation are used to input the teacher model and train our model for execution actions. To maintain the fairness of the experiment, we also replaced diffusion with our VQ-VAE~\citep{razavi2019generating} to provide image input. We do not use 3D structure as input directly. Instead, we use distillation to integrate knowledge. Our simple architecture with one agent and a LLM improves output significantly, surpassing even more complex structures.
\textbf{STEVE-2} can improve the quality by 3.2 $\times$ in FID score and still rank first in the subjective preference scores of GPT-4V(vision)~\citep{yang2023dawn} and human while ensuring accuracy.

\subsection{Ablation Study}
\begin{table}[!t]
\centering
\resizebox{\linewidth}{!}{
\begin{tabular}{@{}l|c|cc|cc|cc@{}}
\toprule
\multirow{2}{*}{\bf Setting} & \multirow{2}{*}{\# LLMs} & \multicolumn{2}{c|}{\bf Goal Search} & \multicolumn{2}{c|}{\bf Block Search} & \multicolumn{2}{c}{\bf Map Exploration}\\
\cmidrule(lr){3-4} \cmidrule(lr){5-6} \cmidrule(lr){7-8}
& & \# iters~($\downarrow$) & success~($\uparrow$) & \# iters~($\downarrow$) & \# blocks~($\uparrow$) & \# iters~($\downarrow$) & area~($\uparrow$)  \\
\midrule
\multirow{2}{*}{w/o KD} & 1 & - & - & - & 8 & 12 & 82 \\
                & 8 / 8 / 8 & 96 & 12\% & 87 & 23 & 6 & 165\\
\midrule
\multirow{2}{*}{w/o EE} & 1 & 41 & 22\% & 35 & 55 & 6 & 172 \\
               & 8 / 8 / 8  & 37 & 45\% & 3 & 312 & 3 & 706 \\ 
\midrule
\multirow{2}{*}{TO} & 4 & 16 & 47\% & 11 & 68 & 4 & 211\\
                        & 24 / 32 / 32 & 5 & 51\% & 2 & 367 & 1 & 1368\\
\midrule
\multirow{2}{*}{\bf STEVE-2} & 1 & 15 & 66\% & 12 & 73 & 6 & 170\\
                        & 5 / 8 / 8 & \textbf{4}  & \textbf{95\%} & \textbf{1} & \textbf{621} & \textbf{1} & \textbf{1821}\\
\bottomrule
\end{tabular}
}
\caption{\textbf{Ablation studies} for multi-modal goal search, continuous block search, and map exploration. The setting is the same as the above experiments. Note that w/o KD is without the knowledge distillation, w/o EE is without the extra expert, and TO is the teacher only with extra knowledge for warm-up.}
\label{tab:ablation_n}
\end{table}
\begin{table}[!t]
\centering
\resizebox{0.8\linewidth}{!}{
\begin{tabular}{@{}l|c|cc|cc@{}}
\toprule
\multirow{2}{*}{\bf Setting} & \multirow{2}{*}{\# LLMs} & \multicolumn{2}{c|}{\bf Material Collection} & \multicolumn{2}{c}{\bf Building Creation}\\
\cmidrule(lr){3-4} \cmidrule(lr){5-6}
& & \# iters~($\downarrow$) & completion~($\uparrow$) & FID~($\downarrow$) & Preferance~($\uparrow$)\\
\midrule
\multirow{2}{*}{w/o KD} & 1 & - & 31\% & - & -\\
                & 8 / 2 & 71 & 52\% & - & -\\
\midrule
\multirow{2}{*}{w/o EE} & 62 & 41 & 71\% & 212.47 & 4.1 \\
               & 8 / 2  & 33 & 77\% & 103.41 & 5.7  \\ 
\midrule
\multirow{2}{*}{TO} & 1 & 67 & 81\% & 125.97 & 4.9 \\
               & 20 / 8  & 32 & 89\% & 87.35 & 5.6 \\ 
\midrule
\multirow{2}{*}{\bf STEVE-2} & 1 & 41 & 95\% & 67.38 & 5.5\\
                        & 8 / 2 & \textbf{4}  & \textbf{99\%} & \textbf{21.12} & \textbf{7.7}\\
\bottomrule
\end{tabular}
}
\caption{\textbf{Ablation studies} for material collection and building creation. The setting is the same as the above experiments}
\label{tab:ablation_c}
\end{table}

As shown in \Cref{tab:ablation_n} and \Cref{tab:ablation_c}, we observe that the  \textbf{STEVE-2} model outperforms the teacher model, GPT-4V(vision)~\citep{yang2023dawn}, significantly in system performance, achieving 1.8 $\times$ better navigation efficiency and 4 $\times$ higher creation quality. Knowledge distillation further enhances the model, improving navigation efficiency by up to 24 $\times$ and enabling the production of high-quality buildings from scratch. For more details of running efficiency, please refer to~\Cref{table:sup_efficiency}.

\vspace{-10pt}

\noindent\paragraph{Hierarchical knowledge distillation is critical.} 
After knowledge distillation, the resulting model can be significantly improved compared to the original, STEVE-13B~\cite{zhao2023see}. Moreover, in the field of creation, knowledge distillation allows the model to produce high-quality buildings from scratch. This method of knowledge distillation enhances multi-agent tasks and provides detailed comprehension of abstract tasks by sorting them out.

\vspace{-10pt}
\noindent\paragraph{Extra knowledge for training is effective.}
With extra expert module, \textbf{STEVE-2} receives additional prior knowledge from the teacher model, leading to high-quality training data and significant task improvement. During testing, we outperform the teacher model without any prior assistance, demonstrating the module's effectiveness in enhancing performance.

\section{Conclusion}
\textbf{STEVE-2} is a novel framework that overcomes limitations of multi-modal language models (MLMs) in open-ended embodied tasks. It uses a hierarchical structure for nuanced task division, a mirrored distillation approach for harnessing parallel simulation data, and an imagination model to infuse extra contextual knowledge into simulations. This boosts the autonomy and effectiveness of embodied agents, bridging the gap between task understanding and execution, and dynamically adapting to open-ended environments. \textbf{STEVE-2} is more sophisticated, flexible, and efficient in complex, real-world applications, advancing the field of artificial intelligence and embodied systems.

\bibliography{main}
\bibliographystyle{colm2024_conference}

\newpage
\section{Appendix}

\paragraph{The supplementary material is structured as follows:}
\begin{itemize}

\item [$\bullet$] First, the ``Implementation Detail'' section outlines the steps involved in implementing STEVE-2, including the hierarchical structure, multi-modal memory and Extra Expert module, in~\cref{sec:implementation}.
\item [$\bullet$] Next, the ``Component Detail' section offers training, inference details and running efficiency comparison, in~\cref{sec:component}
\item [$\bullet$] The ``Case Study'' section presents practical demonstrations of STEVE-2 in creation tasks, supported by figures content, in~\cref{sec:case}.
\end{itemize}

\subsection{Implementation Detail}~\label{sec:implementation}

The Centralized Training with Decentralized Execution~(CTDE)~\citep{hu2023attention,sunehag2017value,lowe2017multi,mao2018modelling} framework for MARL evokes a hierarchical auto-organizing system that leverages global state information for planning while allowing local agents to execute tasks independently. We also utilize a Multi-modal Memory with Retrieval-Augmented Generation~(RAG)~\citep{lewis2020retrieval,mao2020generation} for long-term planning capability and Multi-modal retrieval technique for efficient access to a rich repository of multi-modal memories. The dynamic map visually represents the exploration domain, showcasing only the areas the agents have explored, which is crucial for navigation-like tasks.

\subsubsection{Centralized Planning with Decentralized Execution}\label{sec:CPDE}

We propose Centralized Planning with Decentralized Execution~(CPDE), an architecture for cooperative LLM-based Multi-Agent systems. Based on the Centralized Training with Decentralized Execution~(CTDE)~\citep{hu2023attention,sunehag2017value,lowe2017multi,mao2018modelling} framework for MARL, CPDE leverages global state information for planning while allowing local agents to execute tasks independently. This architecture maximizes global oversight and local autonomy, enabling efficient task completion in complex environments.

\paragraph{Centralized planning for manager agent.}

The centralized planning process for a manager agent $M$ simulates a high-level understanding like that of a human planner. The process includes understanding the environment's dynamic global states, recognizing the conductor agents' capabilities and limitations, and devising a strategy.

\paragraph{Decentralized execution for action agent.}

The decentralized execution process is designed to capitalize on the autonomy and flexibility of conductor agents $C$. These agents navigate the environment, perform tasks, and learn from their interactions guided by the strategic direction from the centralized planning of the manager agent. Note that these conductors can deploy several action agents for similar goals of one sub-goal.

\paragraph{Auto-organizing mechanism.}

We use an auto-organizing mechanism to group agents efficiently. During the planning phase, the manager agent auto-organizes conductor agents based on tasks using MLM and AutoAgents~\citep{chen2023autoagents}. In the execution stage, we use a novel auto-organizing mechanism inspired by Self-Organized Group (SOG)~\citep{shao2022self} for dynamic team composition and varying observability. Each group has a conductor-action agent consensus, and LLM is utilized to summarize~\citep{ma2023large} and distribute messages for a unified schedule.

\subsubsection{Multi-modal Memory}\label{sec:mmm}
Research~\citep{hong2023metagpt} suggests that memory mechanisms are crucial for the functioning of generalist agents. Equipping STEVE-2 with multi-modal memory improves planning accuracy and consistency by leveraging pre-existing knowledge and real-world experiences without requiring additional model updates. Our multi-modal memory system is a key-value memory model with multi-modal keys, storing successfully executed plans. STEVE-2 generates multi-modal queries based on the current task and situations to retrieve relevant memory entries.

\paragraph{Retrival-augmented storage.}
Retrival-augmented storage~(RAS) enables long-term planning capability by Retrieval-Augmented Generation (RAG)~\citep{lewis2020retrieval,mao2020generation}. RAG improves the quality of responses generated by language models with external sources of knowledge to complement the model's internal representation. Instead of external knowledge libraries, we use the collected multi-modal memory as the knowledge library and retrieve interactive experiences as demonstration prompts to improve the planning results.
The formulation is as follows:
\begin{align}
    p(y \mid x) \approx  \sum_{z\in \text{top-k}(p(\cdot \mid x))}p_\eta (z \mid x) \cdot p_\theta (y \mid x,z),
\end{align}
where $x$, $y$, and $z$ represent instructions, plans, and retrieved memory entries. $p_\eta$ and $p_\theta$ denote retrieval and planning models. This retrieval-augmented planning method helps STEVE-2 to ground its internal knowledge efficiently into open-ended environments. It also leverages historical interaction feedback to solve hallucinations within LLMs and produce more accurate plans.

\paragraph{Multi-modal retrieval.}
Multi-modal retrieval~(MMR) enables efficient access to a rich repository of multi-modal memories. This process is initiated with a query containing textual and visual elements. We utilize the manager's Describer $\mathcal{D}_s$ to align this query with the trajectories stored within the multi-modal memory. The Describer converts visual information into a textual format. This textual description serves as an image tag, amalgamating with other textual data or as a textual representation for audio information.

When a retrieval request is made from the multi-modal memory, especially when the information is an amalgamation of image and text, the Describer module $\mathcal{D}_s$ is employed to transcribe the image into text. Subsequently, this description is used to compute the similarity across the multi-modal memory entries. The top-k most similar entries are retrieved for further processing. The formalization of this retrieval process is as follows:
\begin{align}
    \mathcal{R}(q_t, q_v) = \sigma(\mathcal{D}_s(q_v), q_t),
\end{align}
where $q_t$ denotes the textual query, $q_v$ represents the visual query, and $\mathcal{R}$ signifies the retrieval function. The function $\mathcal{S}$ computes the similarity between the multi-modal memory entries and the query, using the textual description provided by $\mathcal{D}_s$.

\subsubsection{Extra Expert}\label{sec:mmm}
Extra Expert mainly covers two complex tasks: Navigation and creation. We develop the dynamic map for Navigation and the generative model for the Creation task.

\paragraph{Dynamic map.}
The dynamic map visually represents the exploration domain, showcasing only the areas the agents have explored. It is the main way for the manager agent to see the global environment. The map will be used in the form of an image. It is updated in real-time with information from all action agents, providing a strategic overview of the environment. This map is instrumental in planning and executing navigation tasks as it reflects the current knowledge and discoveries made by the agent collective.
The following equations can formalize the dynamic map's function:
\begin{equation}
    \mathcal{M}_t = \mathcal{M}_{t-1} \cup \bigcup_{i=1}^{n}S_{i,t}, \qquad S_{i,t} = F(\mathbf{txt}_{i,t}),
\end{equation}
Where \( M_t \) represents the dynamic map at time \( t \), \( S_{i,t} \) is the state information from agent \( i \) at time \( t \), including text data $\mathbf{txt}_{i,t}$ like the place name, special materials. \( F \) is the function that integrates the text data into the map, and \( n \) is the number of active agents. This real-time updating mechanism ensures that the dynamic map remains an accurate and current representation of the exploration field.

\paragraph{Generative model.}
Although several existing works shows diffusion models can also be a good 2D~\citep{guo2024versat2i,chai2023stablevideo,cao2023difffashion,cao2023image} or 3D object~\citep{ouyang2023chasing} and scene~\citep{deng2023citygen} generator, in this paper, we only demonstrate a very basic VQ-VAE based model. Our VQ-VAE takes a channel 4 for RGBA, with a dim of 64. The target channel is the hidden dimension, 256, and the convoluted dim is 16. We train a 3D auto-encoder to generate 3D occupancy based on datasets collected online. Then, we add the text embedding alignment.

\subsection{Component Detail}~\label{sec:component}
All our experiments are on RTX-4090 24G $\times$ 4. Our STEVE-2 is based on STEVE-13b~\citep{zhao2023see} pretrained on LLaMA2-13b~\citep{touvron2023llama}, and we select gpt-4-1106-vision-preview~\citep{yang2023dawn} as the teacher model. Our simulation environment is based on MineDojo~\citep{fan2022minedojo} and uses Mineflayer~\citep{mineflayer} APIs for motor controls.

\paragraph{VQ-VAE efficiency.} During the 2-hour training of the VQ-VAE model, RTX-4090 was utilized for a total of 70,000 epochs. Inference produces a 3D occupancy in an average of 0.07 seconds.

\paragraph{STEVE-2 efficiency.} As shown in~\Cref{table:sup_efficiency}, our method uses the knowledge distillation method to allow one agent and one language model to replace the original results of multiple agents, greatly improving efficiency and resource usage.
\begin{table}[t!]
\centering
\resizebox{0.75\linewidth}{!}{
\begin{tabular}[c]{l|ccc}
\toprule
Method & \# LLMs & \# time~($\downarrow$) & VRAM~($\downarrow$) \\ 
\midrule
\multirow{1}{*}{Voyager}\citep{wang2023voyager}  & 4 & 91.44 s & -\\
\midrule
\multirow{1}{*}{STEVE}\citep{zhao2023see} & 4 & 28.76 s & 68 GB\\
\midrule
\multirow{1}{*}{Creative Agents}\citep{zhang2023creative} & 4 & 91.44 s & -\\
\midrule
\multirow{1}{*}{\textbf{STEVE-2}} & 1 & 9.22 s & 15 GB\\
\bottomrule
\end{tabular}
}
\caption{\textbf{Comparison on language model efficiency.} \# time is for running time per iteration. It calculates the inference time of all language models of an agent. VRAM is about GPU memory usage. The method based on GPT-4 is through API calls, and there is no calculation. 
}
\label{table:sup_efficiency}
\end{table}

\subsection{Case Study}~\label{sec:case}

\begin{figure}[t]
\centering
    \includegraphics[width=\linewidth]{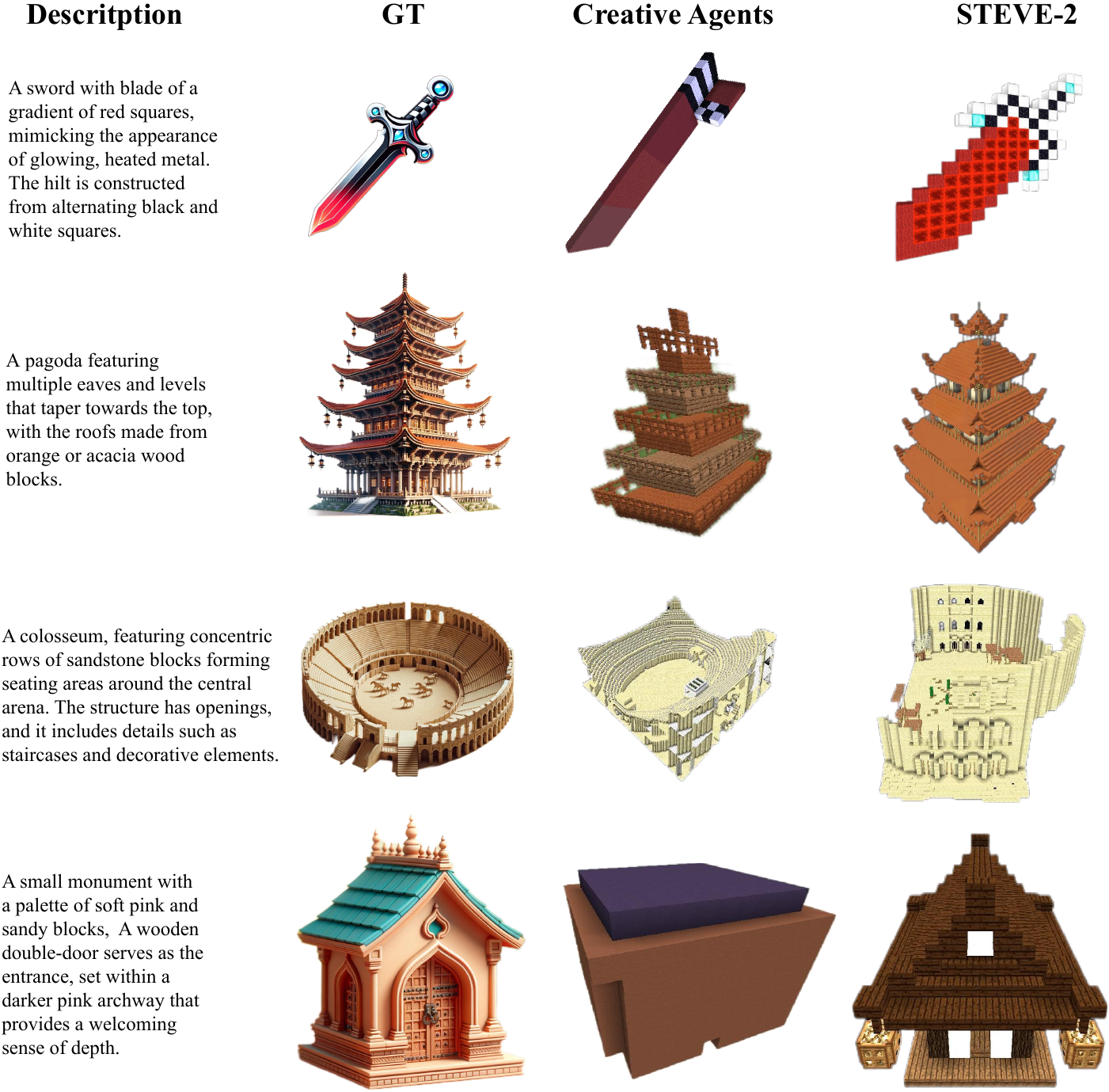}\\
    \caption{\small \textbf{Case study of creation tasks.} GT is the generation output referring to the description by GPT-4 + Dalle3~\citep{betker2023improving}. It is used for FID evaluation of output quality.
    }
    \label{fig:case_creation}
\end{figure}

\begin{figure}[t]
\centering
    \includegraphics[width=\linewidth]{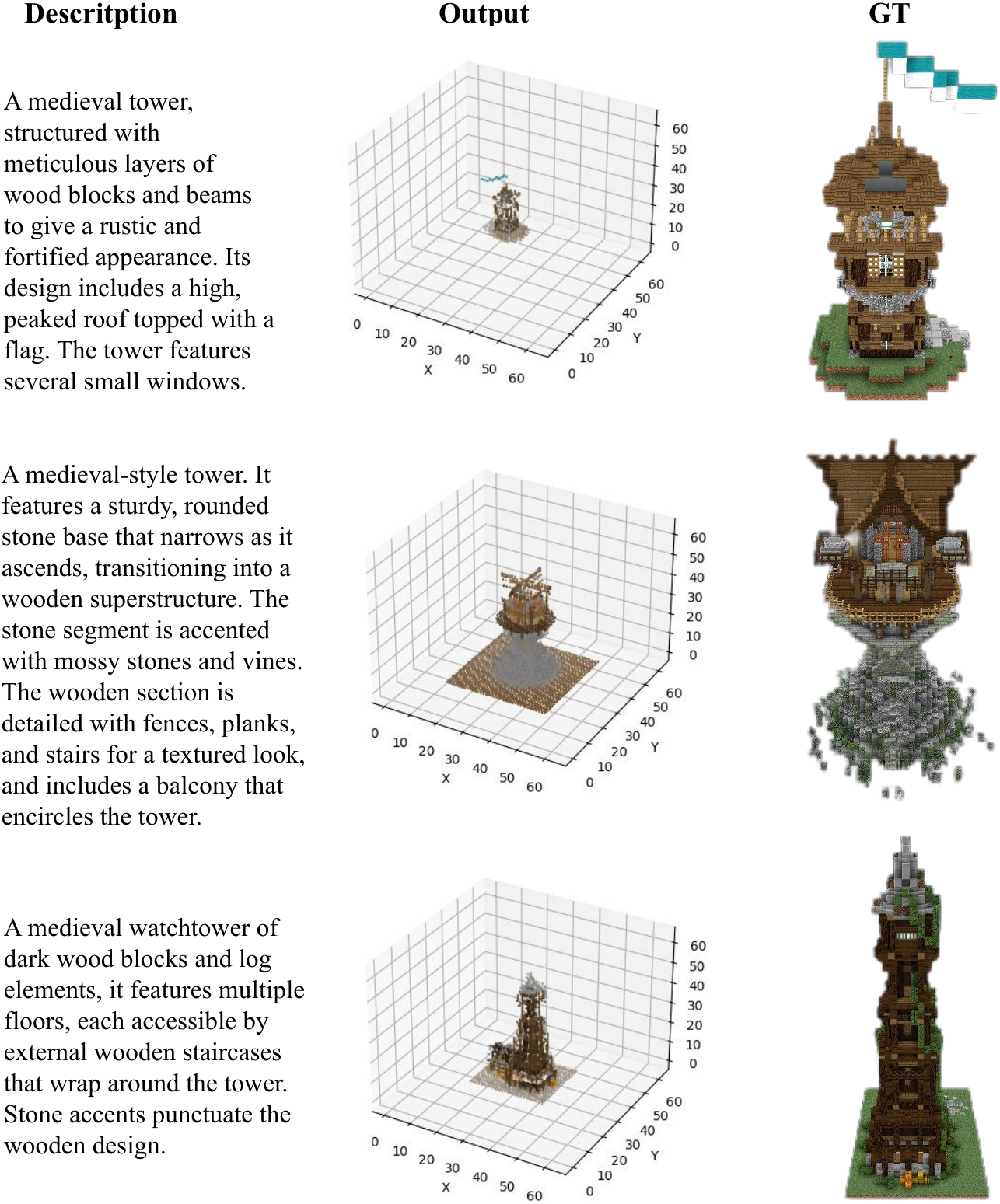}\\
    \caption{\small \textbf{Case study of the generative model.} GT renders 3D occupancy, and Output results from the generative model input through the description.
    }
    \label{fig:case_vae}
\end{figure}

After completing the knowledge utilization process, our work no longer requires any 3D model or picture input to compare reference objects. An additional comparison object must be set up to compare the reference object. As shown in~\Cref{fig:case_creation}, we have listed some representative output results, demonstrating that our method has a more detailed understanding of the preservation of building structures and produces output results more consistent with the description text. However, abstract text can be created in various ways. Therefore, referring to the FID index and using artificial statistical preference is necessary to judge whether the generated results conform to the description. As shown in~\Cref{fig:case_vae}, we also list the generation results of the generative model.

\end{document}